\documentclass{article}

\usepackage[nonatbib,preprint]{neurips_2020}

\usepackage[utf8]{inputenc} 
\usepackage[T1]{fontenc}    
\usepackage{hyperref}       
\usepackage{url}            
\usepackage{booktabs}       
\usepackage{amsfonts}       
\usepackage{nicefrac}       
\usepackage{microtype}      
\usepackage{amsmath}
\usepackage{graphicx}
\DeclareMathOperator*{\argmax}{argmax} 
\usepackage[square,numbers,sort&compress]{natbib}
\usepackage{wrapfig}
\usepackage{subcaption}
\title{Autonomous Control of a Particle Accelerator using Deep Reinforcement Learning}

\author{
  Xiaoying Pang \\
  Apple \\
  \texttt{pangxy@gmail.com}\\
   \and
   \textbf{Sunil Thulasidasan} \\
   Los Alamos National Laboratory \\
   \texttt{sunil@lanl.gov} \\
   \and
   \textbf{Larry Rybarcyk} \\
   Los Alamos National Laboratory \\
  \texttt{lrybarcyk@lanl.gov} \\
}

\begin{document}

\maketitle

\begin{abstract}
We describe an approach to learning optimal control policies for a large, linear particle accelerator  using deep reinforcement learning coupled with a high-fidelity physics engine. The framework consists of an AI controller that  uses deep neural nets for state and action-space representation and learns optimal policies using reward signals that are provided by the physics simulator. For this work, we only focus on controlling a small section of the entire accelerator. Nevertheless, initial results indicate that we can achieve better-than-human level  performance in terms of particle beam current and distribution. The ultimate goal of this line of work is to substantially reduce the tuning time for such facilities by orders of magnitude, and achieve near-autonomous control. 
\end{abstract}

\section{Introduction}

Large particle accelerators are highly complex and dynamic systems. Due to their high-dimensional 
parameter space and dynamic nature, accelerator tuning for optimal performance can be challenging. 
These characteristics, however, makes it potentially suited for a reinforcement learning (RL) approach that 
learns to make sequences of optimal decisions under uncertainty by trial and error. 
Recent developments in reinforcement learning have shown that it can surpass human performance, 
not only in complex board games with exponential search spaces~\cite{silver2017mastering}, but also in real-life settings 
such as controlling large-scale data centers with hundreds of tunable knobs~\cite{evans2016deepmind}, 
finding set points that were previously unknown and unconsidered by human operators. 

Previous research on applying machine learning techniques to accelerator subsystem modeling and 
control have shown promising results \cite{mlacc1, mlacc2}.
In this work, we present an RL approach coupled with deep neural networks that can teach itself
how to operate a section of a particle accelerator just like a human operator does yet without any input 
or supervision from a human. Note that we are not only interested in reaching a good operating condition
from a certain starting point, which can be achieved by many optimization techniques \cite{alex, mopso},
but also if the controller, once trained, can efficiently lead us to the optimal solution from 
virtually anywhere in the parameter space.
 
\section {Reinforcement Learning Background}
Standard reinforcement learning can be framed as a Markov Decision Process (MDP). 
Under this framework, an agent learns to make optimal decisions by interacting with an external environment. 
Learning is an iterative process. At the beginning of the $t$th time step, the agent's status 
quo is captured by a state vector $s_t$. It then takes an action $a_t$ sampled from a 
policy $\pi(a_t|s_t)$, which is a distribution over the actions conditioned on the current 
state $s_t$ of the agent. The policy defines the behaviors of the agent under different 
circumstances. The ultimate goal of a reinforcement learning algorithm is to find the 
optimal policy which will suggest best actions under all possible conditions. After acting out
the action $a_t$, the agent will transition into a new state $s_{t+1}$, according to the transition
probability $p(s_{t+1}|s_t, a_t)$, at the same time, 
observe a consequence or feedback from the external environment,
which is the reward $r_t$. A reward can be positive, indicating some degree of success or negative, which
 indicates a punishment. 
The assumption that the state transition probability to $s_{t+1}$ only depends on the current state $s_t$
and action $a_t$, not on the past ones, makes this process Markov.
The optimal policy we are looking for is not the one that only gives the highest reward at 
the current time step, but the the one that understands delayed gratitude and 
brings the largest total reward eventually. The concept of expected total reward is captured by
two quantities: action-value function $Q^{\pi}(s_t, a_t)$ and
value function $V^{\pi}(s_t)$.

\begin{equation}\label{eq_Q}
Q^{\pi}(s_t, a_t) = E_{\pi}[\sum_{t^\prime = t}^{T} \gamma^{(t^\prime - t)} r(s_{t^\prime}, a_{t^\prime}) | s_t, a_t]
\end{equation}

$Q^{\pi}(s_t, a_t)$ as in (\ref{eq_Q}) is the expected overall reward one can get if 
it starts the $t$th time step at state $s_t$ and takes action $a_t$. 
$\gamma \in (0, 1)$ is a discount factor 
that makes a recent reward more valuable than the future ones and also 
keeps the value function bounded for problems with infinite horizon.
$V^{\pi}(s_t)$ is simply the average of $Q^{\pi}(s_t, a_t)$ over all possible
actions from the policy as shown in (\ref{eq_V}).  

\begin{equation}\label{eq_V}
V^{\pi}(s_t) = E_{a_t \sim \pi(a_t|s_t)}[Q^{\pi}(s_t, a_t)]
\end{equation}

There are several approaches to obtain the optimal policy. One of them is value-based approach 
which adopts a simple greedy policy of taking whatever action that could result in the largest 
Q-function $Q^{\pi}(s, a)$ as shown in (\ref{eq_qlearing}).

\begin{equation}\label{eq_qlearing}
\pi^*(a_t|s_t) = \begin{cases}
    1, & \text{if $a_t = \argmax_a Q^{\pi}(s_t, a)$}.\\
    0, & \text{otherwise}.
\end{cases} 
\end{equation}

The critical part of this approach is to estimate the Q-function and taking the $\argmax$ operation, which could be 
a nontrivial task for continuous action space. Another approach is 
to optimize the policy $\pi_{\theta}(a_t|s_t)$ directly for the best total reward by adjusting its parameter set $\theta$.
This approach is called policy gradient; when combined with an estimated value function, becomes the actor-critic 
approach, with the policy being the actor and the estimated value function being the critic. Both entities can be 
represented by a neural network or any kind of linear or nonlinear model. They are trained simultanously. 
In a forward/prediction step, the best action is predicted by the actor/policy 
and an estimated total future reward is predicted by the critic/value model all based on the current state $s_t$. 
During the backward/update step, the actor/policy is updated to minimize a loss function of  
$-\log\pi_{\theta}(a|s)A^{\pi_{\theta}}(s,a)$
where $A^{\pi_{\theta}}(s,a)$ is the advantage (\ref{eq_adv}). It represents the benefit  
of the total reward by taking a particular action compared to the average one.

\begin{equation}\label{eq_adv}
\begin{aligned}
A^{\pi_{\theta}}(s_t, a_t)  &= Q^{\pi_{\theta}}(s_t, a_t) - V^{\pi_{\theta}}(s_t)\\
Q^{\pi_{\theta}}(s_t, a_t)  &= r(s_t, a_t) + \gamma E_{s_{t+1} \sim p(s_{t+1}|s_t, a_t)}[V^{\pi_{\theta}}(s_{t+1})] \\
				       & \approx 	 r(s_t, a_t) + \gamma V^{\pi_{\theta}}(s_{t+1})
\end{aligned}
\end{equation}

$V^{\pi_{\theta}}(s_t)$ and $V^{\pi_{\theta}}(s_{t+1})$  are estimations provided by the critic/value model which is 
updated by minimizing a mean square loss of $(V^{\pi_{\theta}}(s_t) - Q^{\pi_{\theta}}(s_t, a_t))^2$ 
where $Q^{\pi_{\theta}}(s_t, a_t)$ is obtained according to (\ref{eq_adv}).
More recently, the asynchronous advantage actor-critic (A3C) algorithm \cite{a3c2016} was proposed to enhance the 
performance of the actor-critc approach by asynchronously launching multiple agents in parallel on multiple 
instances of the external environments. 
Since the agents are lauched asynchronously, they are under different circumstances at any given moment,
the consecutive updates from any one of them are more likely to be uncorrelated. This approach can achieve
more stable and efficient training. 
Please refer to the appendix of \cite{a3c2016} for details on the algorithm.

\section{Problem Statement and Method}

Particle accelerators are typically high-capital infrastructures where training an optimal control system using the actual facility is not feasible.
 It is therefore reasonable to 
start training on a high-fidelity simulator that can effectively act as a virtual accelerator
environment. For this purpose, we use a high-fidelity GPU-accelerated
online multi-particle beam dynamics simulator (HPSim~\cite{hpsim}) that simulates a half-mile-long, 
800-MeV proton linear accelerator (linac). The simulator has been successfully used to diagnose and
troubleshoot real-world accelerator operational issues. For this study, the external environment which 
the agent is interacting with is simulated by HPSim and interfaced through OpenAI gym \cite{gym}. 
The goal is to teach the controller to operate a section of the linac (the drift tube linac, or DTL)  by adjusting the cavity field amplitude and phase control variables of the first three tanks. In this case, we have a five-dimensional continuous action space is formed by the 
incremental changes of the cavity field amplitude and phase variables for the first three DTL tanks, 
We chose to work in a continuous action space for convenience  although a discrete action space can also be adopted 
by digitizing the set-point ranges. To avoid the (simulated) tripping of the hardware protection system of an accelerator and making the tuning process more stable, the incremental change step sizes of the action variables are bounded by predefined values as listed in Table \ref{tbl_actionvar}. 
 
\begin{table}[]
\begin{tabular}{llllll}
   &T1 Amplitude & T2 Amplitude & T2 Phase & T3 Amplitude & T3 Phase \\
   \hline
 min &  37.0 & 61.7 &  0.0 & 60.3 &  0.0 \\
 max &  47.6 &  76.1 & 360.0 & 74.2 & 360.0 \\
 max step & 1.0 & 1.0  & 5.0  & 1.0 & 5.0 
\end{tabular}
\caption {Action variable ranges and max adjustment step size.}\label{tbl_actionvar}
\end{table}

We measure the simulated beam current at locations that coincide with the real world counterparts, as well as the power of the lost particles as a proxy for the reading from a real-world beam loss monitor. The beam current and power of lost beam readings at five locations along and after the DTL together with the readings of the absolute values of the five control variables form the 15-dimensional state vector of our RL problem.
A good operating condition is defined as the one when $85\%$ of the beam can reach the last beam
current monitor right before entering the next stage of the accelerator. A learning episode
is considered done once this condition is reached or the number of training steps exceeds a predefined threshold. 
The reward function is defined in (\ref{eq_reward})

\begin{equation}\label{eq_reward}
r = \begin{cases}
    +1000, & \text{if $I[5] / I_0  >= 0.85$}.\\
    -\sum_{n = 1}^{5} [A_n\cdot(0.85 - I[n] / I_0)^2 + B\cdot P[n]^2], & \text{otherwise}.
\end{cases} 
\end{equation}

where $I_0$ is the beam current at the entrance of the DTL and $I[n]$ and $P[n]$ are the beam current 
and power of lost beam reading at the $n$th monitor. One can also make the reward 
function action-dependent 
by subtracting a norm of the action vector to punish extreme changes. In our case, extreme actions are limited
by imposing limits on the sizes of incremental changes of the action variables.  
There are a few hyperparameters in (\ref{eq_reward}).  
\begin{figure}[htbp]
	\centering
	\includegraphics[scale=0.8]{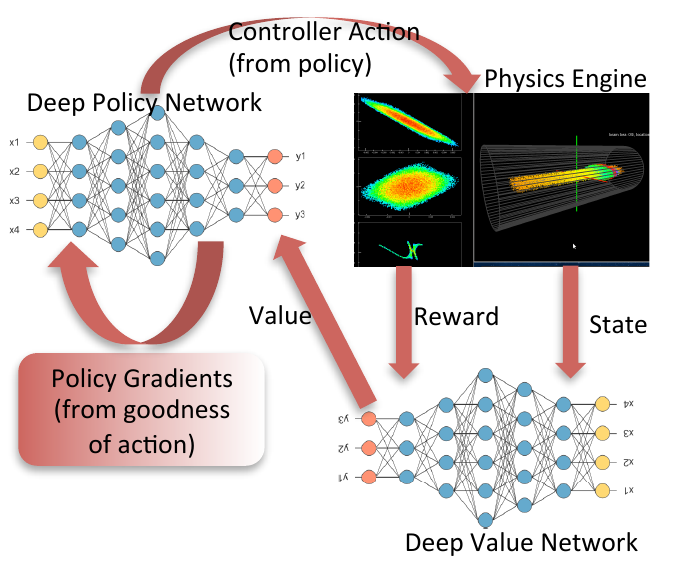}
	\caption{Policy network maps states to actions. The reward and next state are obtained from the simulator. State-space representation and  state values come from a value network; optimal policies are computed via policy gradients}
	\label{fig:controller_schematic}
\end{figure}
We choose to increase $A_n$ as n increases 
to emphasize the importance of 
downstream monitors (A[1] to A[5] range from 0.1 to 0.3 with a step size of 0.05). B is set to be $0.2$.

The learning algorithm we adopt is A3C \cite{a3c2016}. It is efficient and lends itself naturally to problems with continuous
action space. Our simulations are carried out on a cluster node equipped with four Geforce GTX 1080 GPUs.
We can launch up to four agents simultaneously with each running in a separate thread and interacting 
with its own copy of the HPSim accelerator simulator running on one of the GPUs. Each agent also owns a local copy of the RL model which guides its learning locally.
The agents collectively maintain a global RL model by applying their local gradients to it and synchronize their 
local copies with it periodically.
Both the actor and the critic in the RL model are represented as MLP(multilayer perceptron) networks. Figure~\ref{fig:controller_schematic} illustrates the various modules of our learning system.
We used $tanh$ activation function and applied a dropout rate of $10\%$. 
The algorithm predicts a Gaussian policy $\pi_{\theta}(a_t|s_t) = \mathcal{N}(\mu(s_t), \sigma(s_t))$ 
from which the actions are sampled.  The actor/policy loss function is defined in (\ref{eq_loss}).

\begin{equation}\label{eq_loss}
\text{policyLoss} = -\log\pi_{\theta}(a|s)A^{\pi_{\theta}}(s,a) + C_1\cdot\text{actionPenalty}
\end{equation}

The first term in (\ref{eq_loss}) comes from the most commonly used policy gradient estimator \cite{cs294}. However, it leads to an undesirable property that a policy with a positive advantage could end up with a larger loss compared to one with negative advantage. Therefore we tried another loss function with the $log$ in the first term of (\ref{eq_loss}) removed as shown in (\ref{eq_loss2}).

\begin{equation}\label{eq_loss2}
\text{policyLoss} = -\pi_{\theta}(a|s)A^{\pi_{\theta}}(s,a) + C_1\cdot\text{actionPenalty}
\end{equation}

In practice, we were able to achieve better performance by using (\ref{eq_loss2}). However, (\ref{eq_loss2}) alone can still lead to destructively large policy updates. So approaches like trusted region \cite{trpo} and proximal policy optimization algorithms \cite{ppo} should be adopted in the future to ensure better and more stable gradient updates.

The actionPenalty in the second terms of (\ref{eq_loss}) and (\ref{eq_loss2}) is the area under the curve of  $\pi_{\theta}(a|s)$ where the sampled action $a$ can go beyond the limits as listed in Table \ref{tbl_actionvar}. It is nonzero when the policy distribution is likely to predict actions that will step out of the bound.
To encourage exploration, we can include an extra term $-C_2\cdot\text{entropy}$ in the policy loss.
However, in practice, we realized that if we allow $\sigma$ to change and encourage large entropy, we could end up with a broad Gaussian policy which will not be very useful. So we adopted the $\epsilon$-greedy approach instead, by taking the suggested action from the policy with a chance of $(1-\epsilon) 100\%$ and taking a feasible random action otherwise.

\section{Experiments and Results}
\begin{figure}[t]
		\centering
		\includegraphics[scale=0.35]{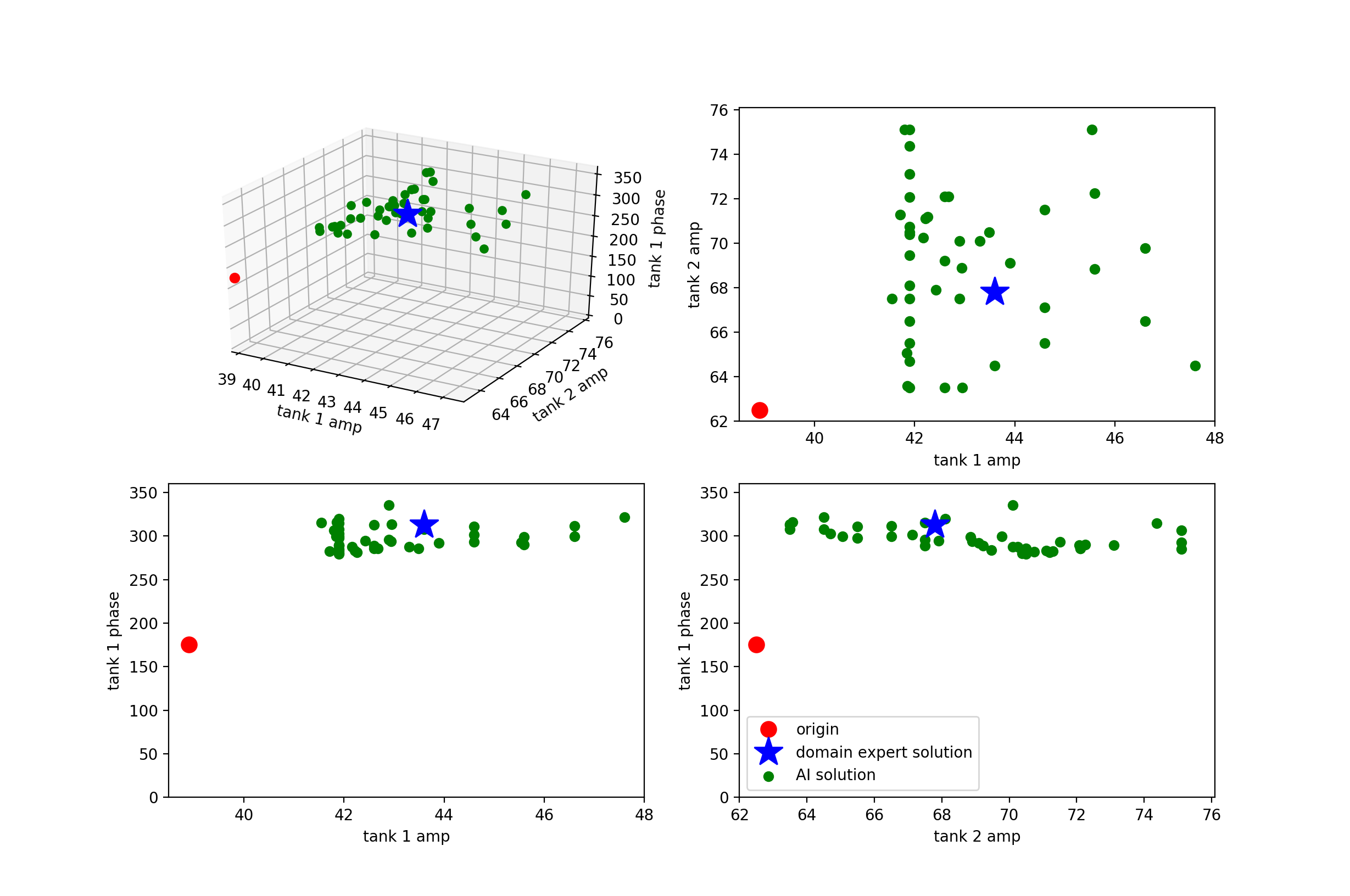}
		\caption{RL solutions in 3D parameter space and their projects onto 2D spaces. Green dots : optimal solution from RL.
			Red dot: starting point of learning. Blue star: domain expert solution. (correction: the axis label 
			`tank 1 phase' should be replaced with `tank 2 phase').}
		\label{fig_solution_space}
\end{figure}

We started the experiment with tuning three action variables, namely the cavity amplitude control variables of the first
two DTL tanks
 and the phase variable of the second tank.
 For this experiment, we launched one agent and used the policy loss function in (\ref{eq_loss2}) and a DNN with one hidden 
layer of 10 nodes to model the policy 
and the value function. One 
 \begin{wrapfigure}[13]{r}{.3\textwidth}
	\includegraphics[scale=0.3]{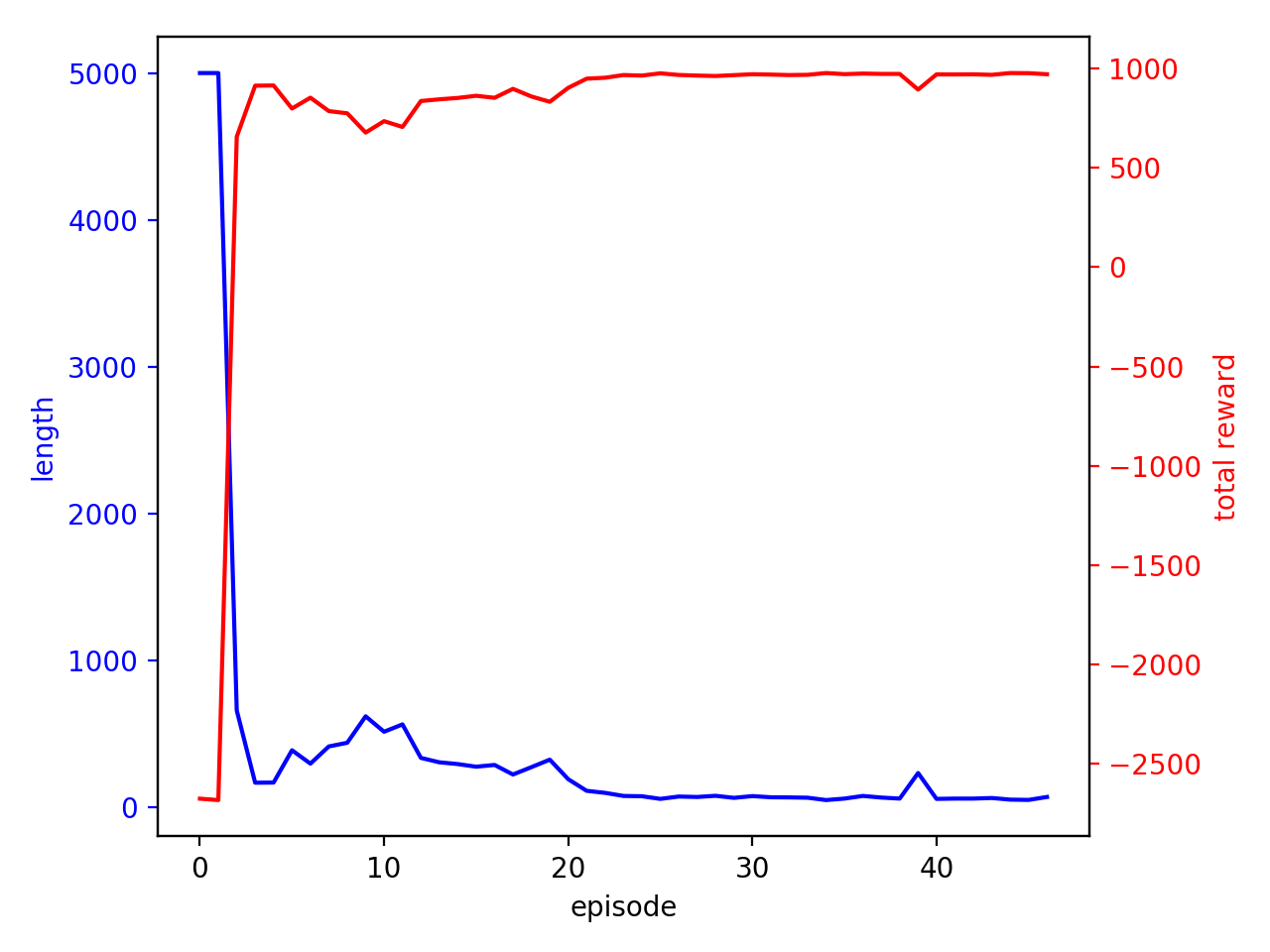}
	\vspace*{-0.2in}
	\caption{Training episode length and collected reward.}
	\label{fig_3action}
\end{wrapfigure}
learning episode ends when 
$85\%$ of the input beam survives or a maximum number
of training steps of 5000 is reached.
We made the agent to return to the same state at the beginning of every episode. 
Figure~\ref{fig_3action} shows how the length and the collected reward of each training 
episode evolves as the training goes on.
After the training is done, the solutions from the learning algorithm is shown in 
Figure~\ref{fig_solution_space}. We can see that the optimal solution is not a single
point in the parameter space. Due to the nonlinearity of the problem, there could be a band of solutions.

\begin{figure*}
  \centering
  \begin{subfigure}[b]{0.45\linewidth}
  \includegraphics[width=\columnwidth]{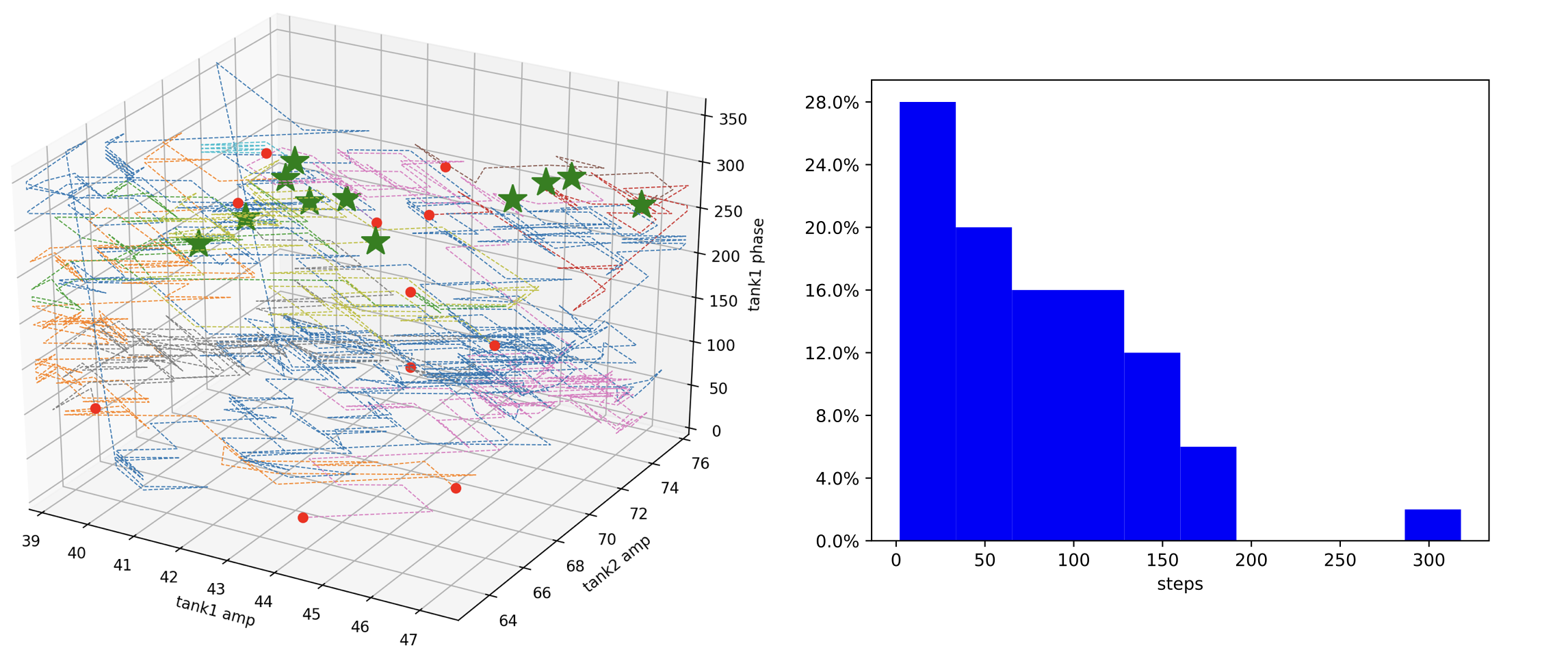}
  \caption{(Left)Learned RL algorithm finds paths to optimal operating conditions (green stars) from 
randomly selected initial points (red dots) in the action parameter space. (Right) Histogram of steps needed to reach optimal condition. Sample size is 50.}
  \label{fig_random_start}
  \end{subfigure}
	\hspace{0.2in}
  \begin{subfigure}[b]{0.45\linewidth}
  \includegraphics[width=\columnwidth]{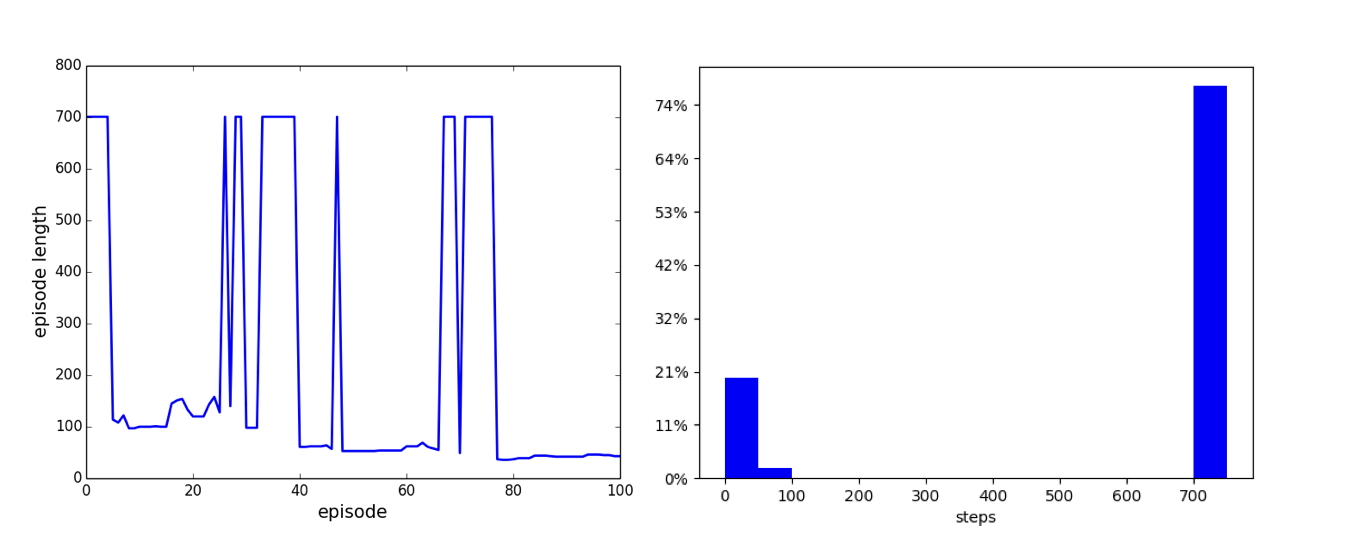}
	\caption{(Left)Training episode length for tuning 5 action variables. Max episode length is 700. (Right): histogram of steps needed to reach optimal solution. Episodes with $steps = 700$ failed to reach optimal solution. Sample size is 200.}
	\label{fig_5act_hist}
\end{subfigure}
\caption{}
\end{figure*}

In Fig. \ref{fig_random_start}, we tested how well the algorithm has learned by 
starting randomly in the state space and tracing where the algorithm directed us.
The learned algorithm can lead us to optimal solutions starting from all the randomly 
selected initial states even if it has only been trained to start from a specific one. 

When we expanded our training to five action variables (including the cavity amplitude and phase variable for the third tank of the DTL), the generalization was much poorer compared to training only three action variables. In Fig.\ref{fig_random_start}, the trained controller was able to start from all the randomly selected initial states and reach a optimal operating condition within 300 steps, whereas in Fig. \ref{fig_5act_hist}), only $21\%$ of the random initial states lead to successful tuning (reaching the goal within 700 steps). The controller was stuck in one of the many local minima for the rest of the cases. This is due to the combined effect of expanding the action variable space and reducing the max number of steps allowed for each learning episode. For this experiment, we used DNNs with two fully connected hidden layers (with 40 and 20 nodes). 


We launched four agents to simultaneously explore the parameter space from four different randomly
selected initial states. An additional agent was created to monitor the progress of the learning by simply executing the policy from a pre-defined starting point.
The maximum number of steps allowed for each episode is reduced to 700 for this experiment.
 In Figure~\ref{fig_5act_hist}, the training episodes are  more unstable than the previous result 
with one agent (Fig. \ref{fig_3action}). This is mainly due to the fact that the  agents now all start 
from completely different initial states. Some of the initial states 
might be closer to the optimal ones than the others. So the best tuning strategy for one of the
exploring agents might not be optimal for the monitoring agent.
As learning increased, we observe the episode length and total reward became more stabilized 
(Figure~\ref{fig_5act_hist}) which indicates better generalization. 
\begin{figure}[h]
  \centering
  \includegraphics[scale=0.35]{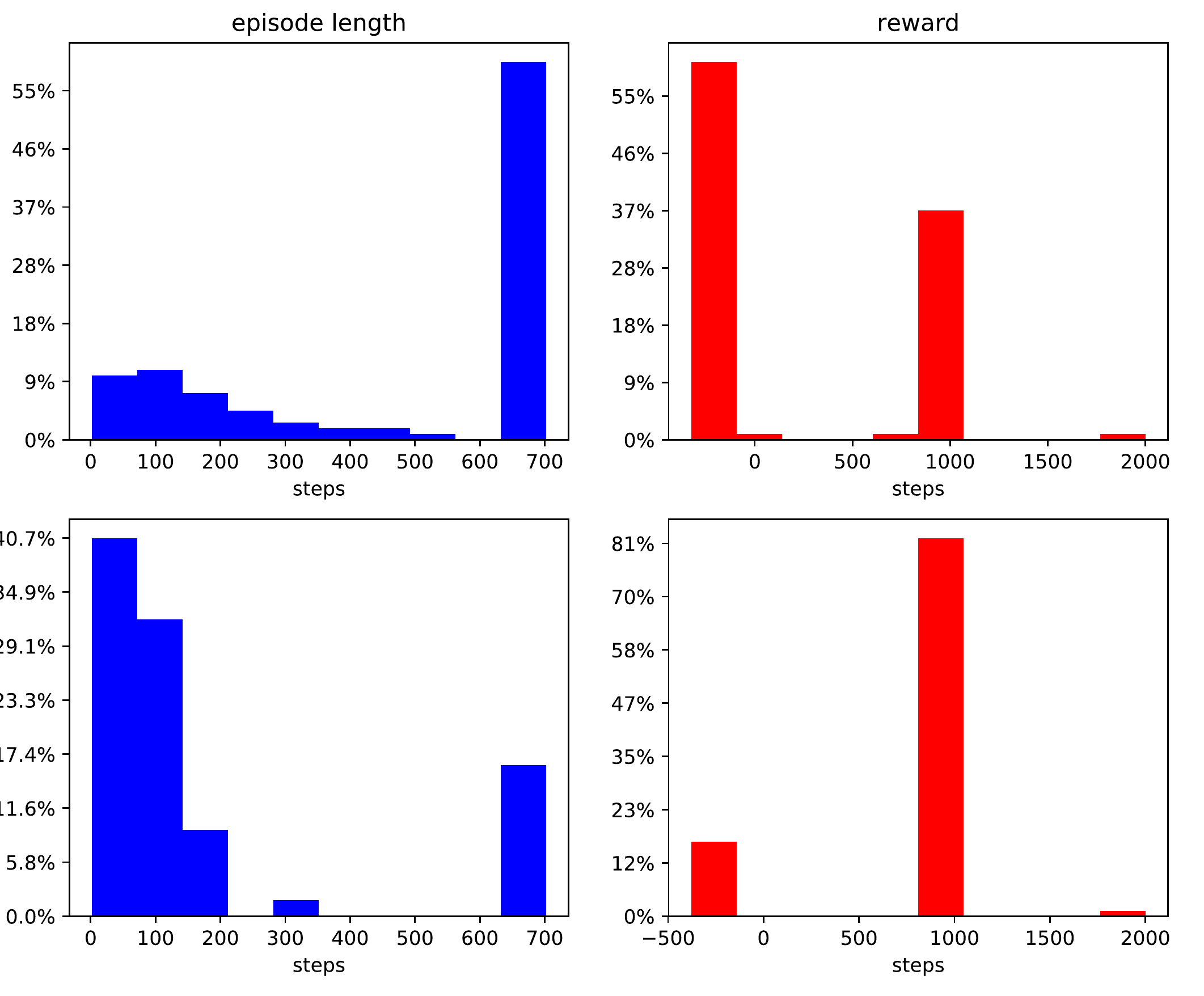}
  \caption{Training episode length and reward of the monitoring agent. Five action variables are tuned.}
  \label{fig_hist}
\end{figure}
We experimented with damping the $\epsilon$ parameter which controls how much exploration 
the RL algorithm is undergoing. We started from a value of $0.5$ and reduced it with each 
episode. 
Fig.\ref{fig_hist} shows the progress made by the RL algorithm by comparing the histogram of
episode length and reward from the first round (top) to that of the final round (bottom). 
In the episode length histograms of Fig.\ref{fig_hist}, the rightmost bar at 700 indicates
the number of unsuccessful episodes. By the end of the second round, the RL controller was
able to steer beam through the DTL with a success rate of $83\%$ and it can 
do that with less than 100 steps about $40\%$ of the times. 

\begin{figure}[h]
  \centering
  \includegraphics[scale=0.5]{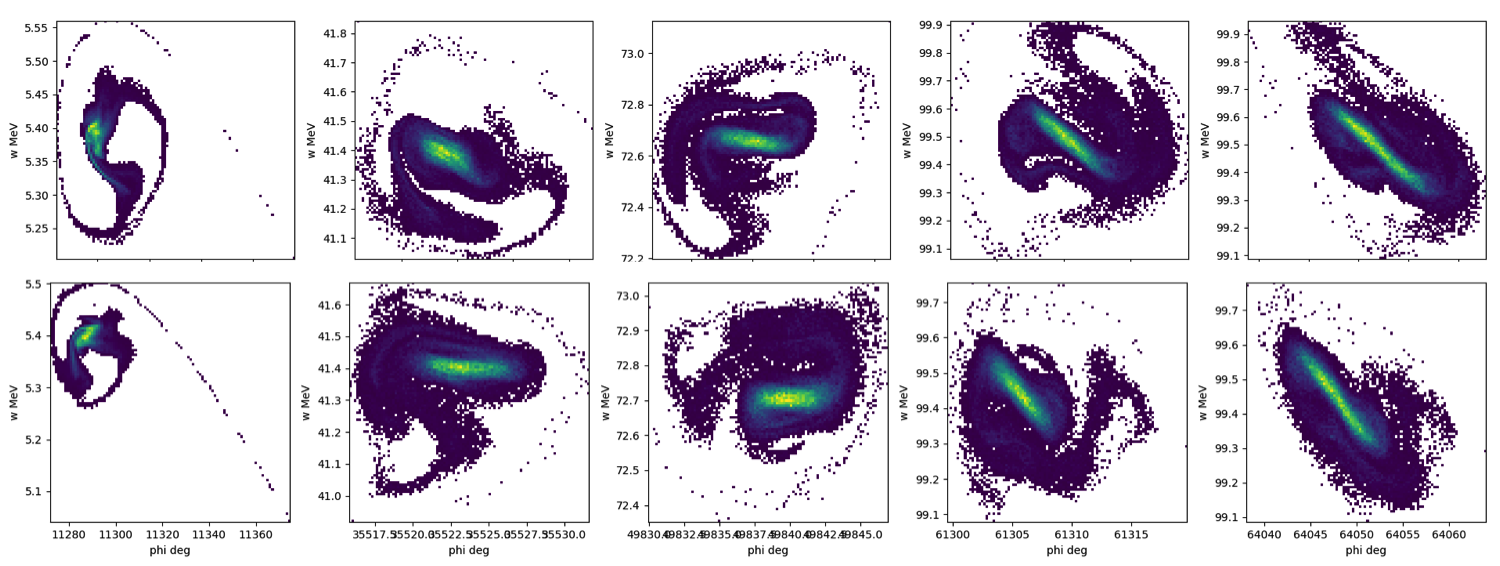}
  \caption{Beam phase-space distributions at 5 different monitoring locations in LANSCE DTL.
The RL solution (bottom) is as good as the domain expert solution (top).}
  \label{fig_phase_space}
\end{figure}

We have also tried to use a combined network for both actor/policy and critic/value function.
It worked reasonably well and achieved similar results but were harder to train and 
took more training rounds to get comparable performance.  Figure~\ref{fig_phase_space} compares the beam phase distributions resulting from the RL training (bottom) to those hand-tuned by a physicist, where we see that the results using our approach look very similar to those of human experts, especially with regard to the size and orientation of the core of the beam (yellow core in the center).


\section{Discussion and Future Work}
In order to learn to control more action variables, we may need to adopt approaches like 
guided policy search \cite{gps} to achieve more efficient training. 
Since the beam passes through each monitor sequentially, it is intuitive to consider breaking up
the learning process into stages and leveraging the result from the previous stage to start the tuning
for the next one. However, we have observed cases where the best operating set-point for a previous
stage of the accelerator ended up being a bad one in order to get good quality beam for the next stage. 
In fact, the observables such as the currents and beam losses can provide only partial information about 
the goodness of a beam in the middle of an accelerator. Other quantities such as beam profile
are equally important. But there is no non-interceptive measurements of these 
quantities available during real-world accelerator operation. Therefore,  when training in a later 
stage, we need to consider action variable from all the stages. Future work includes scaling up this to more action variables and thus a bigger search space, and applying algorithms such as proximal policy optimization~\cite{cs294}.

\bibliographystyle{unsrtnat}
\bibliography{hpsim-a3c}

\end{document}